%% file: root.tex
\documentclass[letterpaper, 10 pt, conference]{ieeeconf}  %

\IEEEoverridecommandlockouts                              %

\usepackage{eso-pic}
\usepackage{url}
\usepackage{microtype}
\usepackage{arydshln} 

\input{includes/includes}
\input{includes/mymacros}

\title{\LARGE \bf
Event-Based Visual Teach-and-Repeat via \\ Fast Fourier-Domain Cross-Correlation 
}

\author{Gokul B. Nair \qquad\qquad Alejandro Fontan \qquad\qquad Michael Milford \qquad\qquad Tobias Fischer %
\thanks{This work was partially supported by funding from ARC DECRA Fellowship DE240100149 to TF, and funding from ARC Laureate Fellowship FL210100156 to MM. The authors acknowledge continued support from the Queensland University of Technology (QUT) through the QUT Centre for Robotics and the Research Engineering Facility (REF) team, especially Dasun Gunasinghe and Joshua Esplin, for providing engineering support, expertise, and research infrastructure. The authors also thank Payam Nourizadeh, Nicolás Marticorena, Vignesh Ramanathan, and Tjeard van Oort for their assistance with outdoor robot experiments.}%
\thanks{The authors are with the QUT Centre for Robotics, Faculty of Engineering, Queensland University of Technology, Brisbane, QLD Australia 4000
{\tt\small gokulbnr@gmail.com}}%
}

\AddToShipoutPicture*{%
     \AtTextUpperLeft{%
         \put(-3.5,10){
           \begin{minipage}{\textwidth}
              \scriptsize
              \MakeUppercase{Preprint version | Paper accepted to the IEEE/RSJ International Conference on Intelligent Robots and Systems (IROS 2026)} 
           \end{minipage}}%
     }%
}

\begin{document}

\bstctlcite{bstctl:forced_etal,bstctl:nodash}

\maketitle
\thispagestyle{empty}
\pagestyle{empty}

\input{text/00abstract}
\input{text/01introduction}
\input{text/02relatedworks}
\input{text/03approach}
\input{text/04experiments}
\input{text/05results}
\input{text/06discussions}
\bibliography{bibfiles/references}
\bibliographystyle{styles/IEEEtran}

\end{document}

%% file: includes/includes.tex
\usepackage[table,dvipsnames,svgnames]{xcolor}
\usepackage{graphicx}
\usepackage{cite}
\usepackage{multirow}
\usepackage{booktabs}
\usepackage{amsmath}
\usepackage{amssymb}
\usepackage{microtype}

\usepackage[utf8]{inputenc}
\usepackage[T1]{fontenc}
\usepackage{hyperref}

%% file: includes/mymacros.tex
\DeclareMathOperator*{\argmax}{arg\,max}

%% file: text/00abstract.tex
\begin{abstract}
Visual teach-and-repeat (VT\&R) navigation enables robots to autonomously traverse previously demonstrated paths using visual feedback. We present a novel event-camera-based VT\&R system. Our system formulates event-stream matching as frequency-domain cross-correlation, transforming spatial convolutions into efficient Fourier-space multiplications. By exploiting the binary structure of event frames and applying image compression techniques, we achieve a processing latency of just 2.60 ms, about 3.5 times faster than conventional camera-based baselines that are optimised for runtime efficiency. Experiments using a Prophesee EVK4 HD event camera mounted on an AgileX Scout Mini robot demonstrate successful autonomous navigation across 3,000+ meters of indoor and outdoor trajectories in daytime and nighttime conditions. Our system maintains Cross-Track Errors (XTE) below 15 cm,  demonstrating the practical viability of event-based perception for real-time VT\&R navigation.
\end{abstract}

%% file: text/01introduction.tex
\section{Introduction}
\label{sec:introduction}

Visual teach-and-repeat (VT\&R) navigation enables robots to autonomously retrace demonstrated paths using visual feedback and is widely deployed from warehouse automation to agricultural robotics~\cite{simon2022performance}. Conventional implementations rely on frame-based cameras to compare current views with stored references and generate corrective control commands~\cite{krawciw2025local}. However, fixed frame rates impose latency between perception and action, constraining achievable update rates and responsiveness.

Event cameras offer a fundamentally different sensing paradigm. Rather than capturing full frames at fixed rates, they asynchronously report pixel-level brightness changes with microsecond temporal resolution~\cite{gallego2020event}. The resulting sparse event stream inherently encodes motion information and reduces redundant processing of static scene regions, thus enabling high-frequency perception-action loops~\cite{gehrig2024low, bauersfeld2026low}. Their unique sensing capabilities provide them with advantages such as a higher dynamic range, minimal motion blur, and low power consumption, making them desirable for onboard sensing capability on energy-limited robotic systems~\cite{hines2025compact, paredes2024fully}. 
\begin{figure}[t]
  \centering
  \includegraphics[width=\linewidth]{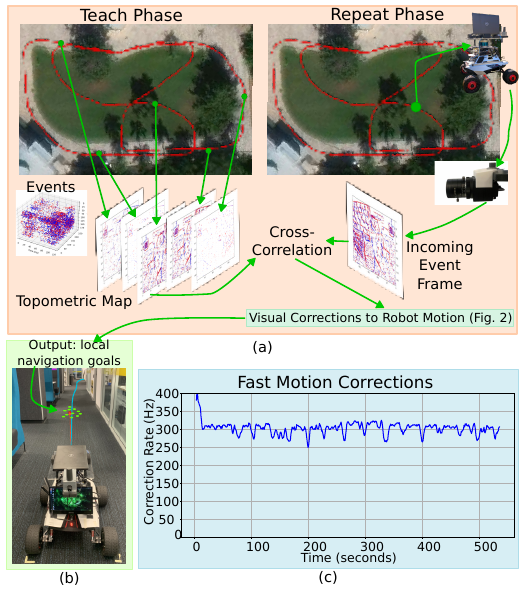}
  \caption{\textbf{Event-based visual teach-and-repeat system overview.} (a): The robot records event streams during the teach phase (left) and autonomously follows the trajectory during repeat (right) using Fast Fourier Transform (FFT)-based cross-correlation of accumulated event frames. (b): Cross-correlations lead to timely navigational corrections in the form of repeatedly updating goals. (c): Our system deployed on an AgileX Scout Mini with Prophesee EVK4 achieves ${\approx}$300 Hz correction rates throughout autonomous navigation.}
  \vspace*{-0.5cm}
  \label{fig:teaser}
\end{figure}

Prior work has explored event-based Simultaneous Localisation and Mapping (SLAM)~\cite{vidal2018ultimate}, visual odometry~\cite{rebecq2016evo}, and energy-efficient drone navigation in simulation~\cite{sanyal2024ev}, but event-based VT\&R has not been demonstrated on real-world ground robots. VT\&R requires efficient reference trajectory storage, real-time matching during repeat traversals, and smooth control generation; requirements that demand algorithms tailored to event data characteristics.

In this paper, we present a novel event-based VT\&R system (Figure~\ref{fig:teaser}). Event streams are accumulated using fixed event counts, which naturally capture more event frames in motion-and-texture-rich areas like corners. During repeat, we perform high-rate matching via efficient frequency-domain correlation. By computing correlations in Fourier space, computational complexity is reduced from $O(N^2)$ to $O(N \log N)$ for an image containing $N$ pixels, enabling adaptive processing rates beyond 300 Hz on consumer hardware.

Extensive experimental validation using a Prophesee EVK4 HD mounted on an AgileX Scout Mini demonstrates successful autonomous navigation across over 3,000 m of indoor and outdoor trajectories. Our system achieves Cross-Track Errors (XTEs) below 15 cm, which is less than that of conventional camera-based baselines~\cite{nourizadeh2024teach, dall2021fast}, while operating at substantially reduced process times.  

Our contributions are threefold:
\begin{enumerate}
    \item \textbf{Event-based VT\&R implementation}: We develop a novel event-based teach-and-repeat system, establishing a baseline for future neuromorphic navigation research and demonstrating the feasibility of event-based trajectory following.
    \item \textbf{High-speed frequency-domain processing}: We introduce a Fast Fourier Transform (FFT)-based correlation framework optimized for the sparse, binary nature of event frames, achieving <3 ms processing time. %
    \item \textbf{Extensive field trials}: We evaluate our pipeline through over 3,000 m of field trials across indoor and outdoor environments. We benchmark against two conventional frame-based VT\&R approaches~\cite{dall2021fast, nourizadeh2024teach} to highlight the performance of our pipeline. 
\end{enumerate}

To support ongoing research in this area, we make the code and the QUT-Event-VTR-Dataset available: https://github.com/gokulbnr/event-vtr. 

%% file: text/02relatedworks.tex
\section{Related Work}
\label{sec:relatedworks}

Visual teach-and-repeat (VT\&R) navigation has evolved from complex metric mapping approaches toward efficient topological methods, while event cameras have emerged as a promising sensing modality for robotic vision. This section reviews teach-and-repeat (T\&R) systems, VT\&R approaches, and event-based perception developments that motivate our work.

\subsection{Teach-and-Repeat (T\&R) Systems}
\label{rel:tnr}

\begin{figure*}[t]
    \centering
    \includegraphics[width=\linewidth]{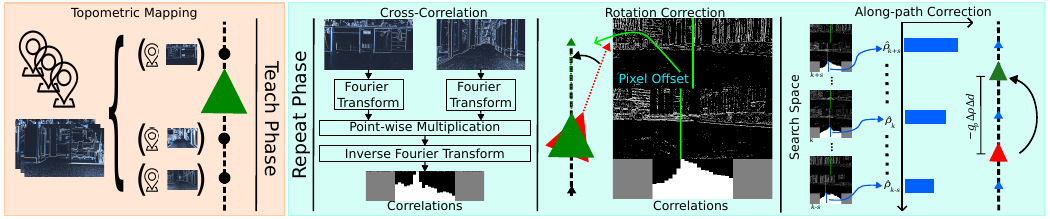}
    \caption{\textbf{Pipeline.} \textbf{Left}: In the teach phase, a topometric map is constructed while teleoperating the mobile robot. The map stores event frames along with corresponding robot poses derived from raw odometry at regular intervals of linear or angular displacement (Section~\ref{method:teach}). \textbf{Left-Centre}: In the repeat phase, as the robot retraces the teach trajectory using the stored poses, incoming event frames are matched with those in the topometric map via cross-correlation. This is performed by point-wise multiplication of image pairs in the Fourier domain (Section~\ref{method:correlation}). \textbf{Right-Centre}: Correlation results yield lateral pixel offsets, which are converted to angular corrections and issued to the robot as updated goal poses (Section~\ref{method:heading}). \textbf{Right}: Along-path corrections are estimated by evaluating correlations across the search space (Section~\ref{method:path}) and applied to the robot’s motion using Equation~\ref{eq:path_corr}. 
}
    \label{fig:corrections}
    \vspace{-0.3cm}
\end{figure*}
T\&R enables mobile robots to autonomously retrace previously recorded trajectories~\cite{simon2022performance}. While GPS provides absolute positioning outdoors~\cite{li1998robot}, its unreliability in indoor and underground environments~\cite{roberts2000autonomous,ruiz2021mental} necessitates alternative sensing approaches. Proprioceptive sensors such as wheel encoders and inertial measurement units (IMU) provide odometry estimates but accumulate drift over time~\cite{nourizadeh2023situ}, requiring exteroceptive sensors for correction.

LiDAR~\cite{sprunk2013lidar} and radar~\cite{qiao2025radar,boxan2025toward} enable robust T\&R through structural matching of point clouds or occupancy grids. However, processing dense 3D data is computationally intensive, limiting deployment on resource-constrained platforms. Early T\&R systems maintained metric maps for precise localization~\cite{krusi2017driving,fox2005hierarchical}, but memory requirements scale poorly with environment size~\cite{aguiar2023topological}. Topometric approaches partially address this through locally consistent submaps~\cite{qiao2025radar,furgale2010visual}, though computational demands remain significant.

Recent work~\cite{nourizadeh2024teach,dall2021fast,truhlavrik2025fast} demonstrates that local correction signals suffice for trajectory following, eliminating the need for metric consistency. These topometric approaches~\cite{krawciw2025local} reduce the teach phase to recording sensor streams alongside odometry, generating corrections through direct sensor comparison during repeat navigation~\cite{nourizadeh2024teach,dall2021fast,krajnik2018navigation}.

\subsection{Visual Teach and Repeat (VT\&R)}
\label{rel:vtnr}

Cameras capture appearance-rich environmental information, enabling place recognition through feature matching across teach and repeat images~\cite{truhlavrik2025fast,simon2022performance}. Traditional approaches extract and match local features~\cite{truhlavrik2025multi}, which is computationally expensive, though efficiency improves by pre-computing and storing teach trajectory features~\cite{krajnik2018navigation,paton2017can}.

Resource constraints on mobile robots have driven development of efficient VT\&R methods~\cite{vcivzek2016low}. For wheeled platforms, heading corrections can be approximated as horizontal image shifts~\cite{dall2021fast,nourizadeh2024teach}. Direct comparison methods include mutual information~\cite{dame2013using,raj2016appearance} and cross-correlation~\cite{matsumoto1996visual}. Normalized cross-correlation (NCC) variants improve robustness~\cite{dall2021fast,nourizadeh2024teach}, while frequency-domain formulations transform convolution into element-wise multiplication~\cite{zhang2009robust}. Our work adapts frequency-domain cross-correlation~\cite{zhang2009robust} for event-based imagery, introducing runtime optimizations specific to event data characteristics. 

\subsection{Event-Based Vision for Robotics}
\label{rel:event}

Event cameras asynchronously detect per-pixel brightness changes, producing events $e_i = (t_i,u_i,v_i,p_i)$ with timestamp $t_i \in \mathbb{R}^+$, pixel coordinates $(u_i,v_i) \in \mathbb{N}^2$, and polarity $p_i \in \{-1,+1\}$ indicating brightness decrease or increase for the $i^{th}$ event~\cite{gallego2020event}. This enables low-latency sensing~\cite{gehrig2024low} and control~\cite{bauersfeld2026low} with minimal power consumption~\cite{sanyal2024ev}, making them suitable for resource-constrained robots~\cite{hines2025compact,paredes2024fully}.

Processing strategies range from per-event~\cite{paredes2024fully} to frame-based accumulation~\cite{rebecq2016evo,fischer2020event,fischer2022many}. High event rates can overwhelm bandwidth, necessitating adaptive filtering~\cite{glover2018controlled} or bias adjustments~\cite{nair2024enhancing,sefidgar2024autobiasing}. Event cameras have been evaluated in autonomous driving~\cite{fischer2022many,pan2025nyc,carmichael2025dataset,hu2020ddd20,gehrig2021dsec}, optical flow~\cite{zhou2021event, li2025emoflow}, drone control~\cite{paredes2024fully,falanga2020dynamic}, power-efficient drone-navigation~\cite{sanyal2024ev} and visual place recognition (VPR)~\cite{hines2025compact}. Concurrent work~\cite{ling2025improving} has explored extracting and matching local features from event frames to perform VT\&R. However, performing cross-correlation on event frames to perform event-driven VT\&R remains unexplored.

Our work bridges this gap by developing the first event-based VT\&R system, which performs cross-correlation on event frames, leveraging event sparsity and high temporal resolution for responsive navigation at high processing rates on constrained platforms.

%% file: text/03approach.tex
\section{Methodology}
\label{sec:approach}

Our system, illustrated in Fig.~\ref{fig:corrections}, uses event count binning to form event frames (Section~\ref{method:event_stream}). During the teach phase, we record event frames and odometry into a topometric map (Section~\ref{method:teach}). In the repeat phase, we match incoming event frames against stored references to correct heading and along-path drift (Section~\ref{method:repeat}). We describe event-frame compression and horizontal concatenation of the search space as means to adhere to strict real-time requirements in Section~\ref{method:fast}.

\subsection{Event Representation}
\label{method:event_stream}

Events $e_i = (t_i, u_i, v_i, p_i)$ are accumulated into binary event-frames $\mathbf{I}_k \in \{-1, 1\}^{H \times W}$ indicating the presence of at least one event at pixel $(u_i, v_i)$, regardless of polarity. The $k^{th}$ frame $\mathbf{I}_k$ is generated using a sliding window $\psi_k$
\begin{equation}
\psi_k = \{i \in \mathbb{Z} \mid kM \le i < kM + N\},
\end{equation}
containing a fixed number of events $N$, which advances by a stride of $M$ events. The $k^{th}$ event-frame is then populated as:
\begin{equation}
\mathbf{I}_k(u,v) =
\begin{cases}
+1, & \exists e_i\ \text{s.t.}\ i \in \psi_k \text{ and } (u,v)=(u_i,v_i) \\
-1, & \text{otherwise}
\end{cases}.
\end{equation}

By discarding event polarity in favor of binary frames, our cross-correlation module treats positive and negative edges identically. This ensures that polarity reversals caused by opposite-direction angular corrections do not alter the event-frame appearance, maintaining matching consistency.

\subsection{Teach Phase}
\label{method:teach}

During the teach phase, the mobile robot is teleoperated along the intended operational path. 
An event frame $\mathbf{I}_k$ and its corresponding odometry pose $\mathbf{T}^W_k \in SE(2)$, 
expressed in the world frame $W$, are recorded at the $k^{\text{th}}$ location whenever the robot 
travels a distance $\Delta d$ or undergoes an angular displacement $\Delta \alpha$. 
Following Dall'Osto et al. \cite{dall2021fast}, we construct a topometric map in the form of an 
ordered list $\mathbf{M}$:
\begin{equation}
    \mathbf{M} = \{(\mathbf{I}_1,\mathbf{T}^W_{1}),(\mathbf{I}_2,\mathbf{T}^W_{2}),\ldots,(\mathbf{I}_K,\mathbf{T}^W_{K})\}.
\end{equation}

\subsection{Repeat Phase}
\label{method:repeat}
During repeat, the robot autonomously follows the teach trajectory using a two-stage motion controller \cite{dall2021fast}. 
The odometry-based controller (Section~\ref{method:lowlevel}) drives the robot to discrete 3-DoF poses derived from $\mathbf{T}^W_{k}$ stored in the topometric map $\mathbf{M}$. 
However, raw odometry is susceptible to cumulative errors, resulting in both lateral and along-path drift \cite{nourizadeh2023situ}. 
To mitigate these effects, cross-correlations are performed in the frequency domain between incoming event frame $\mathbf{\hat{I}}$ and the corresponding teach frames $\mathbf{I}_k$ from the topometric map $\mathbf{M}$ (Section~\ref{method:correlation}) to generate vision-based corrections for lateral alignment (Section~\ref{method:heading}) and along-path position (Section~\ref{method:path}).

\subsubsection{Odometry Driven Motion}
\label{method:lowlevel}
The mobile robot executes the repeat trajectory as a sequence of incremental goal transformations $\mathbf{T}_{k}^\Delta$, defined as:
\begin{equation}
    \mathbf{T}_{k}^\Delta = \left(\hat{\mathbf{T}}^W_{O}\right)^{-1} \, \mathbf{T}^W_{k},
    \quad \forall\, (\mathbf{I}_k,\mathbf{T}^W_{k}) \in \mathbf{M}.
    \label{eqn:goal}
\end{equation}
Here, $\hat{\mathbf{T}}^W_{O} \in SE(2)$ denotes the robot’s current odometry-estimated pose in the world frame $W$.

\subsubsection{Cross-Correlation in Fourier-Domain}
\label{method:correlation}
The incoming node $(\hat{\mathbf{I}}, \hat{\mathbf{T}}^W_O)$, consisting of the latest event frame $\hat{\mathbf{I}}$ and the current odometry-based pose $\hat{\mathbf{T}}^W_O$, is matched against the teach node $(\mathbf{I}_k, \mathbf{T}^W_k) \in \mathbf{M}$, where $k$ denotes the current target pose of the low-level controller. The circumflex notation denotes incoming data during a repeat traverse, differentiating from recorded data in the teach phase. Since odometry accumulates drift \cite{nourizadeh2023situ}, we assume this drift remains bounded within a search window of $s$ frames on either side of $k$, where $s$ is empirically determined based on expected odometry error rates. This assumption defines a search space $\mathcal{S} = \{\mathbf{I}_j : j \in [k-s, k+s]\} \subset \mathbf{M}$, within which candidate nodes are correlated with the incoming event frame $\hat{\mathbf{I}}$.

A cross-correlation score $\mathcal{P}_j \in \mathbb{R}^w$, where $w$ is the image width in pixels, is computed across all possible translational offsets between the incoming event frame $\hat{\mathbf{I}}$ and each event frame in the search space $\mathbf{I}_j \in \mathcal{S}$ as:
\begin{equation}
    \mathcal{P}_j = \mathcal{F}^{-1}\!\left( \mathcal{F}(\mathbf{I}_j) \cdot \mathcal{F}(\hat{\mathbf{I}}^*) \right),
    \quad \forall\, \mathbf{I}_j \in \mathcal{S},
    \label{eq:correlation}
\end{equation}
where $\mathcal{F}(\cdot)$ and $\mathcal{F}^{-1}(\cdot)$ denote the Fourier and inverse Fourier transforms, $\cdot$ denotes point-wise multiplication in the frequency domain, and $^*$ indicates a flip along both image axes.

The cross-correlation step in Equation~\ref{eq:correlation} is restricted to horizontal offsets, based on the assumption that heading errors in wheeled robots with rigidly mounted cameras manifest primarily as lateral shifts \cite{dall2021fast}. To enforce this constraint, images are aligned row-wise, and only the teach-phase event frames $\mathbf{I}_j \in \mathcal{S}$ are zero-padded along the horizontal dimension, yielding a one-dimensional correlation score $\mathcal{P}_j$. The repeat-phase event frames are not padded. The offset $\delta_j$ is then defined as the shift corresponding to the maximum correlation value $\rho_j$:
\begin{equation}
\delta_j = \argmax_{\delta \in [-w/2, w/2]} \mathcal{P}_j(\delta), \quad \rho_j = \max_{\delta \in [-w/2, w/2]} \mathcal{P}_j(\delta).
\label{eqn:max_corr}
\end{equation}
The offsets are measured from the image center, and are clipped at $w/2$. The pixel offset $\delta_j$ corresponding to the maximum similarity is finally converted to a rotational offset as $\theta_j = \frac{\text{FOV}}{w} \delta_j$, where $\text{FOV}$ is the camera’s horizontal angular field of view. 

\subsubsection{Lateral Corrections}
\label{method:heading}

Similar to Dall'Osto et al. \cite{dall2021fast}, heading corrections are interpolated between the previous goal pose $\mathbf{T}^W_{k-1}$ and the current goal pose $\mathbf{T}^W_{k}$, both in world frame $W$, using an interpolation factor $u$, defined as:
\begin{equation}
    u = \frac{\Big((\mathbf{T}^W_{k-1})^{-1} \mathbf{T}^W_{k}\Big)_t \cdot \Big((\mathbf{T}^W_{k-1})^{-1} \mathbf{T}^W_C \Big)_t}
    {\left|\left|\Big((\mathbf{T}^W_{k-1})^{-1} \mathbf{T}^W_{k}\Big)_t\right|\right|^2} ,
    \label{eqn:dist_proportion}
\end{equation}
where $(\cdot)_t$ extracts the translational component of a 3-DoF pose, $||\cdot||$ indicates the Euclidean norm, and $\mathbf{T}^{W}_{C}$ is estimated transform of camera $C$ in world frame $W$. The orientation correction is then computed as:
\begin{equation}
    \Delta \theta = (1-u)\theta_{k-1} + u\theta_k ,
    \label{eq:weighted_angle}
\end{equation}
where $\theta_k$ is the lateral offset estimated from the event frame associated with the current goal $k$, and $\theta_{k-1}$ is the corresponding offset for the previous goal $k-1$. The interpolated lateral offset $\Delta \theta$ is then applied to the mobile robot’s current goal pose $\mathbf{T}_{k}^\Delta$ (from Equation~\ref{eqn:goal}) as:
\begin{equation}
    \widetilde{\mathbf{T}}_{k}^\Delta \leftarrow \mathbf{R}(-g_\theta \Delta \theta) \mathbf{T}_{k}^\Delta,
    \label{eq:rot_corr}
\end{equation}
where $\widetilde{\mathbf{T}}_{k}^\Delta$ is the corrected goal pose, $\mathbf{R}(\cdot) \in \mathrm{SO}(2)$ denotes the planar rotation matrix constructed from the angle $-g_\theta\Delta\theta$, and $g_\theta$ is a calibrated scalar gain parameter.

\subsubsection{Along-Path Corrections}
\label{method:path}

It follows that the correlation values $\rho$ obtained from Equation~\ref{eqn:max_corr} for each event frame in the search space $\mathcal{S}$ attain their maximum for the frame corresponding to the mobile robot’s current location. Following Dall'Osto et al.~\cite{dall2021fast}, noise-level correlations are suppressed by applying a threshold $\bar{\rho}$:
\begin{equation}
    \hat\rho_j=\max(0,\rho_j-\bar\rho),
    \qquad j\in\{k-s,\ldots,k+s\}.
\end{equation}
The along-the-path offset $\Delta \rho$ is computed as a weighted average of the correlations $\hat{\rho}_j$:
\begin{equation}
    \Delta \rho = \frac{\sum_{j=k-s}^{k+s} (j \hat{\rho}_j)}{\sum_{j=k-s}^{k+s} (\hat{\rho}_j)} - u
    \label{eq:weighted_avg}
\end{equation}
where $u$ is the proportion of the distance traveled between the previous and current goal from Equation~\ref{eqn:dist_proportion}. The along-the-path correction $\Delta \rho$ is applied to the mobile robot's current goal as a scalar multiplicative factor:
\begin{equation}
    (\widetilde{\mathbf{T}}_{k}^\Delta)_t \leftarrow \left( \frac{\left\| (\mathbf{T}_{k}^\Delta)_t \right\| - g_\rho \Delta \rho \Delta d}
     {\left\| (\mathbf{T}_{k}^\Delta)_t \right\|} \right) (\mathbf{T}_{k}^\Delta)_t,
     \label{eq:path_corr}
\end{equation}
where $\widetilde{\mathbf{T}}_{k}^\Delta$ is the corrected goal pose, $g_\rho$ is a calibrated scalar gain, and $\Delta d$ is the distance between consecutive goals. %

\subsection{Computational Optimizations}
\label{method:fast}

The real-time requirements of VT\&R necessitate further optimizations beyond frequency-domain processing. We introduce two complementary strategies that exploit the sparse, binary nature of event frames to accelerate computation.

\subsubsection{Event-Frame Compression}
\label{method:compression}

The cross-correlation function is computed as a point-wise matrix product in the Fourier domain, where processing time scales with image dimensions (Equation~\ref{eq:correlation}). We exploit the binary nature of event frames, where most pixels remain inactive (value -1), to enable aggressive compression without significant information loss. Event frames are compressed prior to correlation by applying a one-dimensional summation kernel of size $C_k \times 1$ with stride $C_k$ along each row. This operation reduces the number of columns by a factor of $C_k$, thereby decreasing the computational cost of the subsequent Fourier-domain product.

\subsubsection{Horizontal Concatenation of Search Space}
\label{method:concatenation}

Comparing a repeat frame with each image in the search space from the teach phase traditionally requires multiple forward and inverse transformations between the spatial and Fourier domains. To mitigate this overhead, all teach-phase frames are first concatenated horizontally into a single extended frame. A single Fourier transform is then applied to this combined representation, reducing the number of required transformations while still enabling efficient cross-correlation with the repeat frame. Individual correlation scores are extracted by cropping the resulting map to the original frame width.

%% file: text/04experiments.tex
\section{Experimental setup}
\label{subsec:setup}

We evaluate our event-based VT\&R system across diverse indoor and outdoor environments using a mobile robot equipped with an event camera. This section details our implementation parameters (Section~\ref{setup:implementation}), hardware configuration (Section~\ref{setup:platform}), and experimental scenarios (Section~\ref{setup:exp}). We also introduce evaluation metrics in Section~\ref{exp:metric} and baselines for benchmarking in Section~\ref{exp:baseline}.

\subsection{Implementation Details}
\label{setup:implementation}

Events from the sensor are accumulated into event-frames via an adapted implementation of the OpenEB framework \cite{prophesee2025openeb}. During the repeat phase, a dedicated ROS node generates visual corrections from these frames and publishes updated waypoints as target robot poses. These waypoints are processed by an odometry-driven Sliding Mode Controller (SMC) \cite{nourizadeh2024teach}, which issues drive commands to the mobile robot. Functional parameters are reported in Table~\ref{tab:bias}.

\subsection{Platform}
\label{setup:platform}
\begin{figure}[t]
    \centering
    \includegraphics[width=\linewidth]{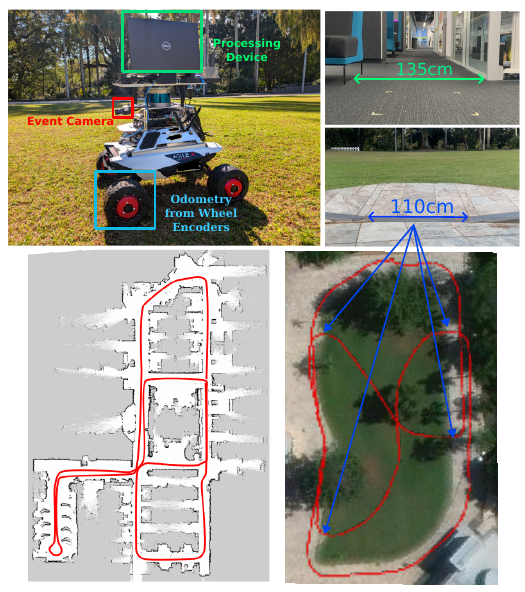}
    \vspace*{-0.15cm}
    \caption{\textbf{Experimental platform and environments.} \textbf{Top-left}: AgileX Scout Mini robot with a front-mounted Prophesee EVK4 HD event camera and an onboard processing laptop. \textbf{Top-right}: Narrow spaces found in our indoor (top) and outdoor (bottom) trial scenarios. \textbf{Bottom-left}: Indoor trajectory visualized on a map from SLAM Toolbox~\cite{macenski2021slam}. \textbf{Bottom-right}: Example outdoor trajectory (223 m) over tiled and grass surfaces.}
    \label{fig:exp}
    \vspace{-0.3cm}
\end{figure}
Our experimental platform consists of an AgileX Robotics Scout Mini robot equipped with a forward-facing Prophesee EVK4 HD event camera~\cite{finateu20205}. Processing is performed on an Intel Core i7-1185G7 connected to the robot via Ethernet. The event camera biases were configured following Pan et al.~\cite{pan2025nyc} to optimize event generation for navigation tasks (Table~\ref{tab:bias}).
Pseudo ground-truth localisation is provided by LiDAR-SLAM both indoors and outdoors~\cite{macenski2021slam}. The complete setup is illustrated in Figure~\ref{fig:exp}.

\subsection{Experimental Scenarios}
\label{setup:exp}

We evaluate our system across two distinct environment types that present complementary challenges for event-based navigation:

\subsubsection{Indoor Workspace} The indoor environment consists of narrow hallways (135 cm wide) lined with repetitive office workspaces, providing only 38.5 cm of lateral clearance for our 58 cm wheelbase robot. These conditions necessitate precise trajectory following despite the challenges of perceptual aliasing from repeating visual patterns. While artificial lighting remains consistent, the varied floor surfaces (carpet and tile) introduce significant odometry drift. Testing trajectories include straight corridors, $90^\circ$ turns, and complete $360^\circ$ rotations (Figure~\ref{fig:exp}), totaling over 1,000 m of indoor evaluation.

\subsubsection{Outdoor University Campus} University campus environments offer longer trajectories (up to 380 m) with greater lateral tolerances but introduce significant natural lighting variations, moving shadows, and varying degrees of surface wetness. Across over 2,000 m of outdoor trials, the robot traverses paved walkways and grass lawns, navigating transition points as narrow as 110 cm. These scenarios pose unique challenges, including texture-poor building walls and wind-induced vegetation motion that can generate spurious events. Such conditions rigorously test the system's ability to maintain stable navigation amid dynamic objects, such as pedestrians and birds, and environmental disturbances that are absent in controlled indoor settings.

\begin{table}[t]
\centering
\caption{Parameters}
\label{tab:bias}
\renewcommand{\arraystretch}{1.1} %
\footnotesize
\begin{tabular}{llc}
\toprule
\textbf{Category} & \textbf{Parameter} & \textbf{Value} \\
\midrule
\multirow{3}{*}{EVK4 Biases} & bias\_diff\_off \& bias\_diff\_on & 40 \\
                             & bias\_fo       & -35 \\
                             & bias\_hpf \& bias\_refr     & 0 \\
\midrule
\multirow{4}{*}{Event Frame} %
                             & Fixed event count, $N$ & $1\times10^5$ events \\
                             & Event window stride, $M$ & $1\times10^3$ events \\
                             & Initial frame dimensions & 1280$\times$720 \\
                             & Downsampled frame dimensions & 320$\times$180 \\
\midrule
\multirow{2}{*}{Teach}       & Distance step, $\Delta d $ & 20 cm \\
                             & Angle step, $\Delta \alpha$ & 15$^\circ$ \\
\midrule
\multirow{4}{*}{Repeat}      & Search space, $\pm s$ & 4 event frames \\
                             & Lateral correction gain, $g_\theta$ & $1.5\times10^{-3}$ \\
                             & Along path correction gain, $g_\rho$ & $1.5\times10^{-5}$ \\
                             & Compression kernel size, $C_k$ & 8 pixels \\
\bottomrule
\end{tabular}
\vspace*{-0.25cm}
\end{table}
\subsection{Metrics}
\label{exp:metric}

In field trials of VT\&R systems, a key performance metric is the success rate (SR), which is calculated as the proportion of repeat trajectories that are completed relative to the total number of repeat experiments conducted. A repeat trajectory is considered a failure if the robot can no longer progress safely, such as when proximity to static obstacles would result in a collision. 

To further analyze repeat trajectories, we compute the Cross-Track Error (XTE)~\cite{baril2022kilometer} between the teach run and subsequent repeats using ground-truth robot positions. The XTE is defined as the distance between the robot's frame at each repeat sample and its orthogonal projection onto the teach path. As in Baril et al.~\cite{baril2022kilometer}, we approximate this orthogonal projection by identifying the spatially closest point on the teach trajectory for each repeat sample. The Euclidean distance between these nearest neighbours is then calculated to obtain the error samples
\begin{equation}
d_i = \min_{j} \left\lVert \mathbf{p}^{\,\text{teach}}_{j} - \mathbf{p}^{\,\text{repeat}}_{i} \right\rVert,
\end{equation}
where $\mathbf{p}^{\,\text{teach}}_{j}$ denotes the robot pose recorded by the ground truth at the $j$-th sample of the teach trajectory, and $\mathbf{p}^{\,\text{repeat}}_{i}$ denotes the $i$-th pose in the repeat trajectory, matched to its nearest neighbour in the teach trajectory.

\begin{table*}[t]
\centering
\caption{Comparison of navigation performance across indoor and outdoor trajectories for our approach (Ours), an odometry-only baseline, and RGB-based VT\&R~\cite{dall2021fast,nourizadeh2024teach}. For each trajectory, the results are reported as the mean and standard deviation of the Cross-Track Error (XTE), in centimetres, for up to three repeated trials, along with the corresponding success rate (SR). The odometry-only results additionally include the travelled distance as a percentage of the full trajectory length (\% Length). Our system achieves comparable performance with the RGB-based VT\&R approaches~\cite{dall2021fast,nourizadeh2024teach}. Our method consistently completes the full trajectories in all repeats, achieving 100\% success rate.}
\label{table:combined}
\renewcommand{\arraystretch}{1.2}
\setlength{\tabcolsep}{2.8pt} %
\small %
\resizebox{\textwidth}{!}{
\begin{tabular}{@{}ll c ccc c ccc c ccc c ccc@{}}
\toprule
& \textbf{Tr.} & \textbf{Len.} 
& \multicolumn{3}{c}{\textbf{Odom-only}} & \multicolumn{4}{c}{\textbf{Dall'Osto et al. \cite{dall2021fast}}} & \multicolumn{4}{c}{\textbf{Nourizadeh et al. \cite{nourizadeh2024teach}}} & \multicolumn{4}{c}{\textbf{Event VT\&R (Ours)}} \\
\cmidrule(lr){4-6} \cmidrule(lr){7-10} \cmidrule(lr){11-14} \cmidrule(lr){15-18}
& \textbf{\#} & \textbf{(m)} & \textbf{\#1} & \textbf{\%L} & \textbf{SR} & \textbf{\#1} & \textbf{\#2} & \textbf{\#3} & \textbf{SR} & \textbf{\#1} & \textbf{\#2} & \textbf{\#3} & \textbf{SR} & \textbf{\#1} & \textbf{\#2} & \textbf{\#3} & \textbf{SR} \\
\midrule
\multirow{3}{*}{\rotatebox[origin=c]{90}{\textbf{Indoor}}} 
& 1 & 65 & 9.85$\pm$13.94 & 17 & 0/1 & 6.08$\pm$6.06 & 6.40$\pm$6.24 & 6.91$\pm$6.80 & 3/3 & 4.58$\pm$3.17 & 4.80$\pm$3.82 & 4.14$\pm$4.03 & 3/3 & 6.96$\pm$6.83 & 6.36$\pm$5.54 & 5.68$\pm$4.83 & 3/3 \\
& 2 & 100 & 6.98$\pm$10.84 & 14  & 0/1 & 6.53$\pm$6.38 & 12.88$\pm$12.07 & 6.89$\pm$5.98 & 3/3 & 9.84$\pm$10.95 & 8.05$\pm$5.18 & 8.98$\pm$8.11 & 3/3 & 8.69$\pm$7.52 & 7.68$\pm$7.88 & 8.82$\pm$9.25 & 3/3\\
& 3 & 200 & 7.03$\pm$9.19 & 5  & 0/1 & 7.51$\pm$7.31 & 11.79$\pm$12.45 & 6.93$\pm$7.13 & 3/3 & 12.93$\pm$11.14 & 11.40$\pm$10.11 & 9.67$\pm$9.15 & 3/3 & 10.94$\pm$8.83 & 8.57$\pm$6.63 & 8.66$\pm$7.32 & 3/3\\
\midrule
\parbox[t]{2mm}{\multirow{3}{*}{\rotatebox[origin=c]{90}{\textbf{Outd.}}}} 
& 4 & 120 & 36.76$\pm$40.55 & 16 & 0/1 & 11.73$\pm$8.74 & 8.61$\pm$6.67 & 31.03$\pm$30.67 & 3/3 & 8.44$\pm$7.02 & 5.23$\pm$4.96 & 12.01$\pm$9.76 & 3/3 & 5.27$\pm$6.48 & 9.99$\pm$6.91 & 8.11$\pm$7.37 & 3/3 \\
& 5 & 223 & 80.03$\pm$110.29 & 19 & 0/1 & 17.67$\pm$16.20 & 20.19$\pm$19.52 & 14.25$\pm$19.63 & 3/3 & 15.62$\pm$16.84 & 17.31$\pm$19.77 & 18.06$\pm$17.92 & 3/3 & 5.78$\pm$5.25 & 11.74$\pm$9.67 & 14.68$\pm$10.72 & 3/3 \\
& 6* & 381 & 27.41$\pm$36.50 & 14 & 0/1 & 13.46$\pm$15.79 & 21.94$\pm$22.84 & 10.81$\pm$12.16 & 3/3 & 9.47$\pm$10.88 & 9.26$\pm$10.22 & 5.80$\pm$6.36 & 3/3 & 14.83$\pm$16.97 & 13.17$\pm$15.57 & 5.22$\pm$3.68 & 3/3 \\
\bottomrule
\multicolumn{18}{l}{\footnotesize \textsuperscript{*} Night-time trajectory.}
\vspace{-0.5cm}
\end{tabular}
}
\end{table*}
\begin{figure*}[t]
    \centering
    \includegraphics[width=\linewidth]{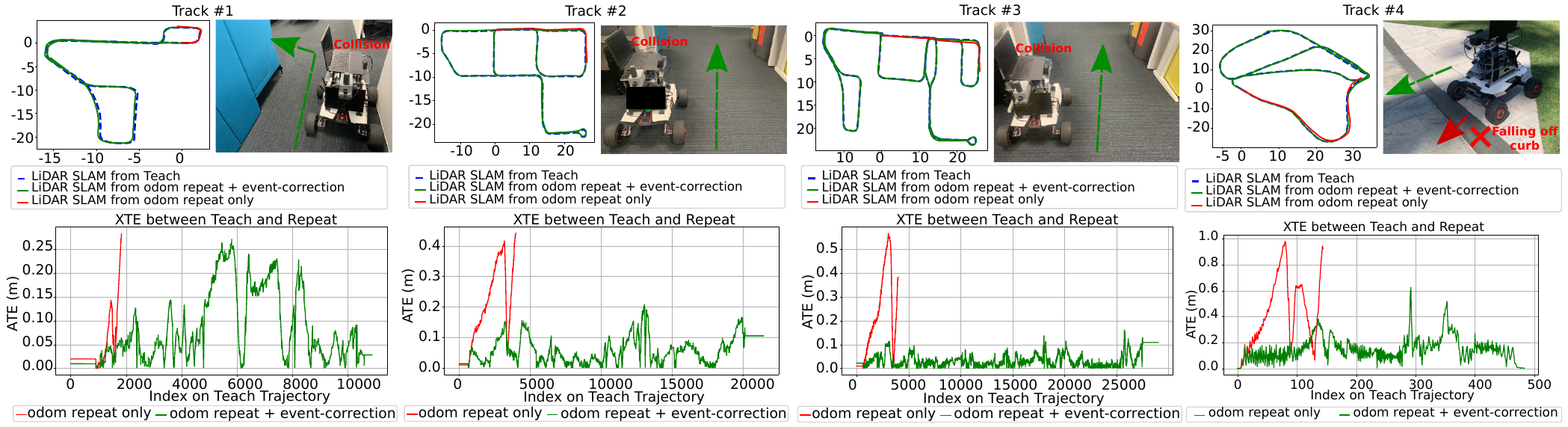}
    \caption{\textbf{Navigation performance across indoor and outdoor trajectories.} \textbf{Top row:} Three indoor trajectories (left) and one outdoor trajectory (right). The \textcolor{blue}{blue} paths denote the teach trajectories, the \textcolor{DarkGreen}{green} paths show the repeat trajectories using our event-based correction, and the \textcolor{red}{red} paths indicate the odometry-only repeats. Both indoor and outdoor trajectories were estimated using LiDAR SLAM. Next to each map, failure cases of the odometry-only baseline are shown. In these examples, the \textcolor{DarkGreen}{green} arrows indicate the direction the robot is expected to traverse for a successful repeat. \textbf{Bottom row:} Cross-Track Error (XTE) for our method (\textcolor{DarkGreen}{green}) and the odometry-only baseline (\textcolor{red}{red}). Note that the odometry-only XTE grows unbounded, making the system prone to collisions and eventual failure.}
    \label{fig:gps_outdoors}
    \vspace{-0.5cm}
\end{figure*}
\subsection{Baselines}
\label{exp:baseline}

To establish benchmarks and demonstrate the effectiveness of our system relative to existing VT\&R approaches, we consider the following baselines:

\noindent \textit{Odom-only Baseline}: This approach relies solely on odometry-driven motion control (Section~\ref{method:lowlevel}) without any visual corrections. This naive baseline allows us to quantify the contribution of our event-driven visual corrections.

\noindent \textit{Dall'Osto et al.}~\cite{dall2021fast}: A conventional camera based VT\&R approach designed for fast image matching using normalized cross-correlation (NCC) and performs repeat traverses using a proportional controller~\cite{corke2011robotics}.

\noindent \textit{Nourizadeh et al.}~\cite{nourizadeh2024teach}: A conventional camera based VT\&R approach which performs image matching using the same normalised cross-correlation (NCC) approach as Dall'Osto et al.~\cite{dall2021fast}, and uses a sliding mode controller (SMC) for odometry-driven robot motion. 

%% file: text/05results.tex
\section{Results}
\label{sec:results}

We evaluate our event-based VT\&R system through extensive field trials covering over 3,000 meters of indoor and outdoor trajectories. Our experiments demonstrate both the navigation accuracy and computational speed enabled by event-camera perception.

\subsection{Navigation Performance}
\label{res:trials}

We evaluate our event-based VT\&R system through two weeks of field trials across the environments described in Section~\ref{setup:exp}, covering six distinct trajectories: three indoor tracks (65-200 m) and three outdoor tracks (120-380 m). 

Table~\ref{table:combined} summarizes the navigation performance across all trials. Our event-based VT\&R system achieved a 100\% success rate (18/18 trials), whereas the odometry-only baseline failed consistently before accomplishing 5–19\% of the full trajectory length (Figure~\ref{fig:gps_outdoors}). The XTE (Section~\ref{exp:metric}) remained consistently low for our event-based system, with average errors of 8.04 cm and 9.87 cm in indoor and outdoor trials, respectively. %

Our system achieves performance on par with conventional camera-based baselines. The method by Dall'Osto et al.~\cite{dall2021fast} yielded 7.99 cm and 16.63 cm, while Nourizadeh et al.~\cite{nourizadeh2024teach} recorded 8.26 cm and 11.24 cm, respectively. Crucially, our event-based VT\&R approach maintains a 100\% success rate (3/3) in challenging night-time conditions, with a mean XTE of 11.07 cm. These results demonstrate that our event-based framework maintains competitive accuracy while offering the significant computational advantages detailed in the next Section~\ref{res:efficiency}.

\subsection{Computational Speed}
\label{res:efficiency}

A key challenge in developing our VT\&R system was designing a cross-correlation function capable of processing high-rate event frames, as detailed in Section~\ref{method:fast}. We benchmark our computational performance against efficiency-focused conventional frame-based approaches~\cite{nourizadeh2024teach, dall2021fast}.

Our pre-processing, which thresholds event pixels into a polarity-agnostic binary event-frame (Section~\ref{method:event_stream}), requires only $0.26$ ms. In contrast, the patch-normalization used by Dall'Osto et al.~\cite{dall2021fast} and Nourizadeh et al.~\cite{nourizadeh2024teach} adds $7.52$ ms to the computational latency. Furthermore, while their systems perform normalized cross-correlation (NCC) to match frames taking $13.31$ ms, our frequency-domain cross-correlation matching records a run-time of only $2.34$ ms.

\subsection{Ablation Studies}
\label{subsec:ablations}

\subsubsection{Impact of Event-Accumulation Strategies on Velocity Invariance}
\label{abl1:velocity_invariance}

\begin{figure}[t]
    \centering
    \includegraphics[width=\linewidth]{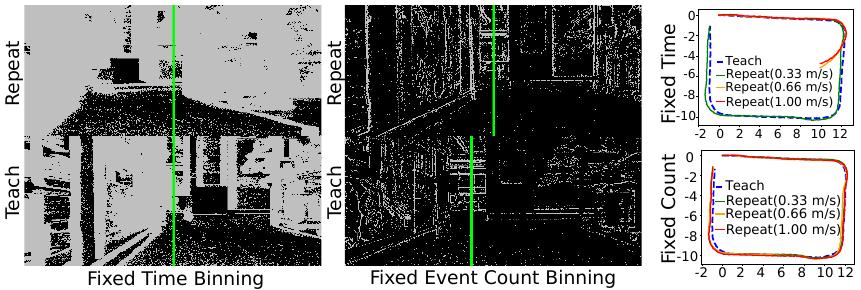}
    \caption{\textbf{Event-Accumulation Strategies under Varying Linear Velocities.} (Section~\ref{abl1:velocity_invariance}) \textbf{Left}: Comparison of Repeat and corresponding Teach frames using fixed-time binning. Significant appearance divergence occurs during angular motion, leading to navigation failure.
\textbf{Center}: Comparison of frames using fixed-event count binning (proposed) at the same location. The representations remain consistent despite the change in linear velocity (0.33 m/s vs.~1.00 m/s). 
\textbf{Top-Right}: When using fixed-time accumulation, repeats at $0.66$ and $1.00$ m/s fail for teach traverse taken at $0.33$ m/s.
\textbf{Bottom-Right}: For our proposed fixed-event count strategy, all three repeat traverses are completed with 100\% success rate.}
    \label{fig:abl1}
    \vspace{-0.3cm}
\end{figure}

Our hypothesis is that accumulating events by event counts generates event frames which are more robust to velocity variations when compared to accumulating events by fixed time windows. To verify this, we compare two event accumulation strategies by recording a single teach traverse and performing multiple repeat traverses at varying velocity profiles. Specifically, we record the teach traverse at $0.33$ m/s and repeat traverses across $0.33$, $0.66$, and $1.00$ m/s indoors, whereas for outdoor experiments we record the teach traverse at $1.37$ m/s and repeat the traverse at velocities of $1.00$, $1.50$, and $1.68$ m/s. We record 100\% success rates for experiments with a fixed event count strategy.  Figure~\ref{fig:abl1} shows that fixed time-based event accumulation fails consistently after the first corner in the map when the teach velocity differs from the repeat velocity (Figure~\ref{fig:abl1} Top Right), while our proposed event count based method successfully repeats all traverses, even under large velocity differences between teach and repeat traverses. 

\subsubsection{Impact of Computational Optimizations}
\label{abl2:optim}

To understand the impact of computational optimizations performed in the pipeline (Section~\ref{method:fast}), we perform time-profiling on our pipeline with and without the event-frame compression (Section~\ref{method:compression}) and horizontal concatenation of teach frames for single cross-correlation step across the search space (Section~\ref{method:concatenation}). As reported in Table~\ref{tab:abl2},  horizontal concatenation reduces latency by approximately 36\%, while event-frame compression produces a larger reduction of approximately 86\%.  
\begin{table}[t]
\centering
\caption{Computational optimisations for image matching}
\label{tab:abl2}
\renewcommand{\arraystretch}{1.1}
\footnotesize
\begin{tabular}{cccc}
\toprule
\textbf{Image Compression} & \textbf{Horizontal Concatenation} & \textbf{Time (ms)} \\
\midrule
$\times$ & $\times$ & 26.90 \\
$\checkmark$ & $\times$ & 3.63 \\
$\times$ & $\checkmark$ & 17.19 \\
$\checkmark$ & $\checkmark$ & 2.34 \\
\bottomrule
\end{tabular}
\end{table}

%% file: text/06discussions.tex
\section{Discussion and Conclusions}
\label{sec:discussions}

This paper presents a novel event-camera-based VT\&R system, leveraging the high temporal resolution and asynchronous output of event cameras to enable trajectory corrections at low processing latencies. Event frames are obtained by aggregating the incoming event stream into overlapping event-count windows and are correlated against pre-collected event frames in the Fourier domain. Through extensive field trials in both indoor and outdoor scenarios, we showcase a robust VT\&R system capable of maintaining accurate trajectory following, with robustness showcased through repeat traverses at multiple velocity profiles for a single teach recording. To facilitate future research, we collect and compile raw event data from the event camera, along with ground-truth robot poses from LiDAR SLAM.

This work prioritizes high-frequency perception and control over explicit 3D reconstruction. While repeated trials demonstrate task-completion despite the presence of birds, humans, and swaying foliage, future work could further enhance the robustness of event-based VT\&R in crowded or highly dynamic environments. Incorporating a coarse understanding of the 3D environment derived from auxiliary sensors could enhance system robustness. Furthermore, multi-modal sensing could be integrated through early fusion to refine along-track visual corrections, or via late fusion by arbitrating motion commands between independent event-based pipelines. While we mitigated the motion-dependence of raw events by accumulating them by event count, future research could explore motion-compensation techniques or local feature extraction to ensure visual corrections are generated at a more regular rate. The trials taken to evaluate our pipeline were conducted over two weeks. A limitation of our approach is the long-term repeatability through changes in the appearance of the place being revisited due to day-night shifts, weather changes, seasonal changes, etc.  Future work could explore adapting to such variations between the corresponding teach and repeat traverses. 